%% file: main.tex
\documentclass[journal]{IEEEtran}

\usepackage{times}
\usepackage{graphicx} 
\usepackage{subfigure} 
\usepackage{setspace}

\usepackage{helvet}
\usepackage{courier}
\usepackage{hyperref}

\usepackage{makecell}
\usepackage{graphicx}
\usepackage{subfigure}
\usepackage{color}
\usepackage{amsfonts}
\usepackage{amsmath}
\usepackage{amssymb}

\usepackage{multirow}
\usepackage{booktabs}
\usepackage{cleveref}

\DeclareMathAlphabet\mathbfcal{OMS}{cmsy}{b}{n}

\def\0{{\bf 0}}
\def\1{{\bf 1}}

\usepackage{ntheorem}

\newtheorem*{*thm}{Theorem}

\newtheorem*{*lemma}{Lemma}

\usepackage{multicol}

\usepackage{subfigure}
\usepackage{multirow}
\usepackage{url}
\usepackage{cite}
\usepackage{hyperref}
\usepackage{microtype}
\usepackage{bm}
\usepackage{amssymb}
\usepackage[T1]{fontenc}
\usepackage{colortbl}
\usepackage{booktabs,multirow}
\usepackage{makecell}
\usepackage{tabulary}
\usepackage{fontawesome5}

\newlength\savewidth

\usepackage[utf8]{inputenc} 
\usepackage[T1]{fontenc}    
\usepackage{microtype}      
\usepackage{capt-of}
\usepackage{xcolor}
\usepackage{xspace}
\usepackage{amsmath,amssymb,amsfonts,dsfont,pifont,bm,mathrsfs,mathtools,nicefrac}

\usepackage{booktabs,multirow,adjustbox,diagbox,threeparttable}
\definecolor{citeblue}{RGB}{48,111,186}

\usepackage{graphicx}
\usepackage{makecell}
\usepackage{enumitem}
\usepackage{color, colortbl}
\usepackage{tikz}
\usepackage{subcaption}

\usepackage{algorithm}
\usepackage{listings}
\usepackage{dsfont}
\usepackage{setspace}

\definecolor{tabhighlight}{HTML}{e5e5e5}
\definecolor{natural}{HTML}{648FFF}
\definecolor{specialized}{HTML}{DC267F}
\definecolor{structured}{HTML}{362682}
\definecolor{vtabmean}{HTML}{FE6100}
\definecolor{vtabparam}{HTML}{FE6100}
\definecolor{baselinecolor}{gray}{.9}

\usepackage{graphicx}  
\usepackage{caption}   
\usepackage{float}     

\usepackage{times}
\usepackage{latexsym}

\usepackage[T1]{fontenc}

\usepackage[utf8]{inputenc}

\usepackage{microtype}

\usepackage{inconsolata}
\usepackage{algpseudocode}

\usepackage{graphicx}

\usepackage{color}
\usepackage{listings}

\usepackage{booktabs}

\usepackage{tabularx}
\usepackage{multirow}
\usepackage{amsmath}
\usepackage[table]{xcolor}
\usepackage{amsfonts}
\usepackage{float} 
\usepackage{colortbl}
\usepackage{setspace}
\usepackage[accsupp]{axessibility}  
\title{Sparse-Tuning: Adapting Vision Transformers with Efficient Fine-tuning and Inference}

\author{Ting Liu, Xuyang Liu, Liangtao Shi, Zunnan Xu, Yue Hu, Siteng Huang, Yi Xin, \\ Bineng Zhong, Donglin Wang, \textit{Member, IEEE}

	\IEEEcompsocitemizethanks{
	\IEEEcompsocthanksitem{Ting Liu and Xuyang Liu contributed equally to this work. This work was supported in part by the National Natural Science Foundation of China (Grant Nos. 62306329 and 62103425), the Natural Science Fund of Hunan Province (Grant Nos. 2023JJ40676 and 2022JJ40559) \textit{(Corresponding authors: Yue Hu and Siteng Huang.)}

    Ting Liu and Yue Hu are with the College of Systems Engineering and State Key Laboratory of Digital Intelligent Modeling and Simulation, National University of Defense Technology, Changsha 410072, China. (email: \{liuting20, huyue11\}@nudt.edu.cn). 
    
    Xuyang Liu is with the College of Electronics and Information Engineering, Sichuan University, Chengdu 610065, China (email: liuxuyang@stu.scu.edu.cn). Liangtao Shi is with the Key Laboratory of Knowledge Engineering with Big Data, Hefei University of Technology, Hefei 230009, China, and also with the Ministry of Education and School of Computer Science and Information Engineering, Hefei University of Technology, Hefei 230009, China (e-mail: shilt@mail.hfut.edu.cn).

    Siteng Huang is with the College of Computer Science and Technology, Zhejiang University, Hangzhou 310058, China (email: siteng.huang@gmail.com). Donglin Wang are with the School of Engineering, Westlake University, Hangzhou 310030, China (email: wangdonglin@westlake.edu.cn). Zunnan Xu is with Tsinghua University, Beijing, 100084, China (e-mail: xzn23@mails.tsinghua.edu.cn). Yi Xin is with the State Key Laboratory for Novel Software Technology, Nanjing University, Nanjing 210000, China (email: xinyi@smail.nju.edu.cn).

    Bineng Zhong is with the Key Laboratory of Education Blockchain and Intelligent Technology, Ministry of Education, Guangxi Normal University, Guilin 541004, China, and the Guangxi Key Lab of Multi-Source Information Mining and Security, Guangxi Normal University, Guilin 541004, China (e-mail: bnzhong@gxnu.edu.cn).

}
    
	}
    }

\begin{document}
\maketitle

\input{sec/0_abstract}

\input{sec/1_intro}

\input{sec/2_related}

\input{sec/3_method}

\input{sec/4_exp}
\input{sec/5_conclusion}

\bibliographystyle{IEEEtran}
\bibliography{references}

\end{document}

%% file: sec/0_abstract.tex
\begin{figure*}[t]
    \centering
    \includegraphics[width=\textwidth]{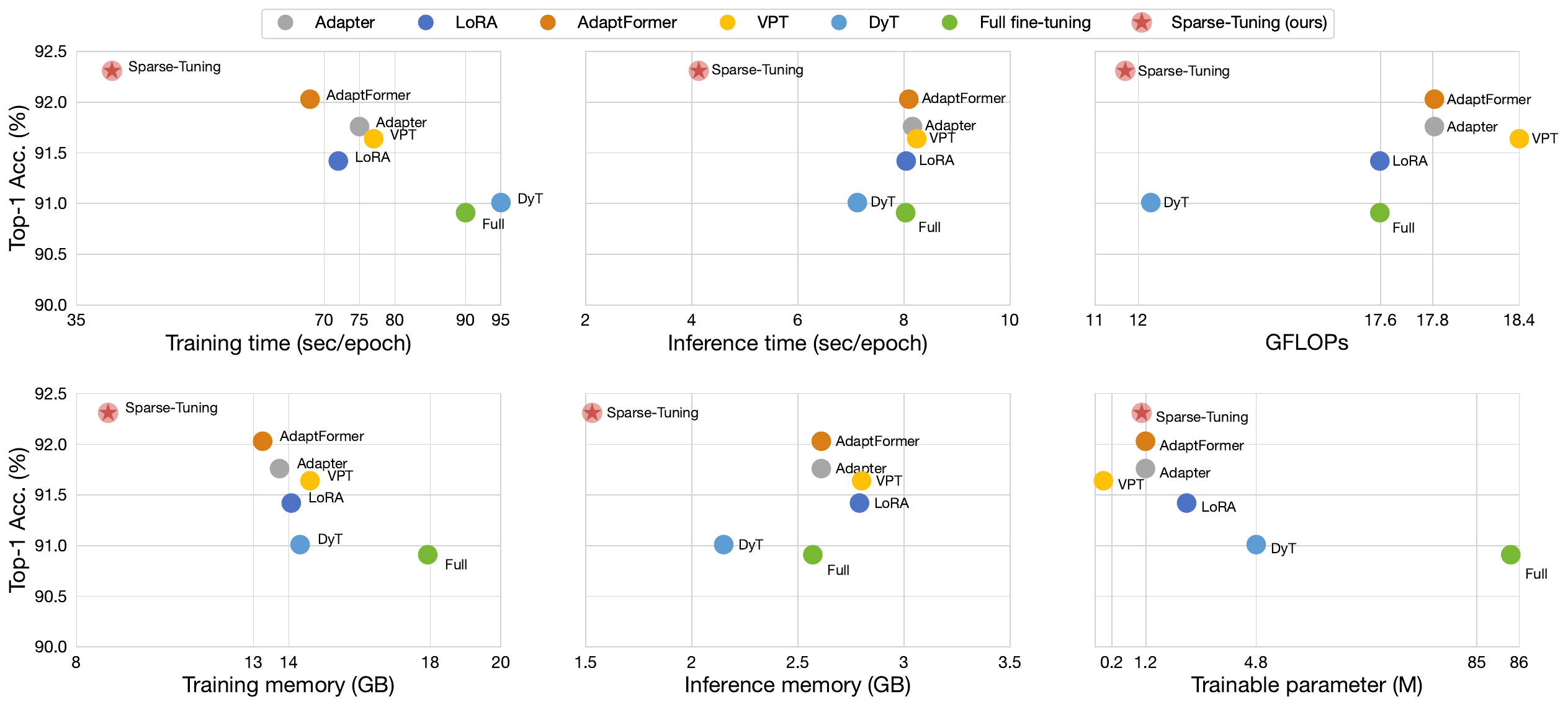}
   
    \caption{\textbf{Comparisons of 
    Sparse-Tuning with other mainstream PEFT methods on CIFAR-100 dataset.} Sparse-Tuning enhances performance while remarkably reducing training and inference time, GPU memory consumption, and computational complexity, achieving both fine-tuning and inference efficiency of the pre-trained ViT.}
  
\label{fig:efficiency}
\end{figure*}

\begin{abstract}
Parameter-efficient fine-tuning (PEFT) has emerged as a popular solution for adapting pre-trained Vision Transformer (ViT) models to downstream applications by updating only a small subset of parameters. While current PEFT methods have achieved fine-tuning efficiency, they overlook the efficiency of computation and GPU memory during inference, falling short of practical requirements. To address this limitation, we propose Sparse-Tuning, an efficient and effective framework that leverages popular token sparsification (TS) techniques to reduce information redundancy in images and videos, thereby significantly improving computational and memory efficiency. However, TS often compromises performance due to inevitable information loss. To address this limitation, we further introduce Dense Adapters (DA) to compensate for the information losses incurred by token sparsification. DA integrates comprehensive token information from shallow layers into the retained tokens of deeper layers, ensuring minimal performance degradation. Through the integration of TS techniques and DA, Sparse-Tuning achieves a significant reduction in computation and memory overhead while maintaining performance. Empirical results on VTAB-1K, three image datasets, and two video datasets show that Sparse-Tuning reduces GFLOPs to 66\% of the original ViT-B while achieving state-of-the-art performance compared to full fine-tuning and other PEFT baselines.

\end{abstract}


%% file: sec/1_intro.tex
\section{Introduction}

Large-scale Vision Transformer (ViT) models~\cite{dosovitskiy2020vit,liu2021swin,radford2021clip,kirillov2023sam} have demonstrated remarkable generalization capabilities across a wide range of downstream vision tasks. The predominant approach adapts these models to specific tasks through the \textit{pretrain-then-finetune} paradigm, where models are initially pre-trained on large-scale datasets~\cite{deng2009imagenet,lin2014coco} and subsequently fine-tuned for individual downstream tasks. However, as these pre-trained ViT models continue to scale up~\cite{zhai2022vit-g,dehghani2023vit-22b}, full fine-tuning becomes increasingly computationally prohibitive. 
Additionally, there are risks of catastrophic forgetting and overfitting when fine-tuning on limited downstream datasets~\cite{zhang2022noah,Diao2023UniPT,liu2024dara}. 
Recently, various Parameter-Efficient Fine-Tuning (PEFT) methods~\cite{houlsby2019adapter,lester2021prompt,hu2021lora,jia2022vpt,chen2022adaptformer,liu2025m2ist,shi2025mamba} have been proposed to mitigate the high fine-tuning costs by updating only a minimal subset of parameters. However, these PEFT methods still \textit{incur substantial computational overhead during inference}.

To alleviate computational challenges during inference, various token sparsification methods (TS) have been proposed to reduce redundant tokens while preserving essential information~\cite{liu2025shifting}. Prominent examples include DynamicViT~\cite{rao2021dynamicvit}, which introduces a prediction module to dynamically identify and prune uninformative tokens, and EViT~\cite{liang2022evit}, which uses the attention scores between the \texttt{[CLS]} token and image tokens to selectively retain the most relevant ones, thus reducing computational overhead. However, directly applying these TS techniques disrupts the original attention distribution, as shown in Figure~\ref{attention} (b). An intuitive approach is to integrate PEFT with TS, achieving efficient adaptation. As shown in Figures~\ref{attention} (c) and (d), although PEFT helps mitigate these issues, it still \textit{fails to fully restore the attention distribution}. Specifically, TS alters the regions the ViT attends to, leading to some performance degradation. This may be because the remaining tokens in deeper layers lose significant spatial information.


To better maintain performance after token sparsification, we propose \textbf{Sparse-Tuning}, a novel PEFT paradigm that achieves efficient fine-tuning and inference with token sparsification mechanism, and also preserves strong performance for ViT adaption. As our ablation study (Table \ref{Table:TS and DA}) indicates, token sparsification leads to inevitable \textit{information loss} and decreased accuracy. After token dropping, the remaining vision tokens in the deeper layers can be considered sparse representations, retaining only coarse and holistic features, whereas the complete tokens in the shallower layers capture the full spatial information of the image. Inspired by this, we meticulously design Dense Adapters (DA) to supplement the local detail information for compressed tokens. Different from conventional intra-layer adapters, our Dense Adapters establish dense connections between different layers, integrating multi-level features from shallower layers to enhance the current tokens. Additionally, the full utilization of local features also contributes to the precise identification of semantic-relevant tokens within the subsequent layers.
With these non-trivial designs, Sparse-Tuning improves performance while significantly reducing computational cost for efficient ViT fine-tuning and inference, as shown in Figure \ref{fig:efficiency}.



To fully evaluate the generalization, we conduct extensive experiments on the common PEFT benchmark VTAB-1K \cite{zhai2019vtab}, three full-scale image datasets, \textit{i.e.}, CIFAR-100 \cite{krizhevsky2009cifar-100}, SVHN \cite{goodfellow2013svhn}, and Food-101 \cite{bossard2014food-100}, as well as two complete video datasets, \textit{i.e.}, Kinetics-400 (K400) \cite{carreira2017k400} and Something-Something V2 (SSv2) \cite{goyal2017ssv2}. Empirical results on VTAB-1K demonstrate that with only \textbf{11.65} GFLOPs and approximately \textbf{66\%} of the computational cost of the original ViT-B, Sparse-Tuning outperforms all state-of-the-art methods in performance and efficiency. Moreover, Sparse-Tuning achieves superior performance in both image and video recognition on complete datasets while significantly improving both fine-tuning and inference efficiency. 

The remainder of this paper is structured as follows. Section~\ref{Related Work} provides a comprehensive review of existing literature on token sparsification techniques for Vision Transformers and parameter-efficient fine-tuning methods. Section~\ref{Method} presents a detailed exposition of our proposed Sparse-Tuning framework. Section~\ref{Experiments} demonstrates the effectiveness of our approach through extensive empirical evaluations, encompassing both quantitative benchmarks and qualitative analyses across diverse image and video tasks. Finally, Section~\ref{Conclusion} summarizes our work and discuss the limitations and future work in Section~\ref{Future_Work}.

%% file: sec/2_related.tex
\section{Related Work}
\label{Related Work}

\subsection{Token Sparsification for ViT}

Recent works have explored to accelerate the inference efficiency of ViT \cite{rao2021dynamicvit,liang2022evit,Bolya2023ToMe,li2024high,wang2025divico}, with most of them aiming to reduce the token redundancy to decrease computational complexity. For instance, a series of studies have focused on token sparsification. Specifically, DynamicViT \cite{rao2021dynamicvit} prunes uninformative tokens identified through prediction modules. Following a similar adaptive principle, DVT \cite{wang2021dvt} enhances computational efficiency by automatically determining the optimal token count for each input image. Meanwhile, SuperViT \cite{lin2023supervit} handles diverse patch sizes with a single model, adaptively adjusting token retention during inference. In contrast to pruning, another line of research employs token merging. For example, EViT~\cite{liang2022evit} retains the top-k most attentive tokens based on their attention scores with the \texttt{[CLS]} token and merges the rest via weighted averaging. Similarly, ToMe~\cite{Bolya2023ToMe} utilizes Bipartite Soft Matching (BSM) to compute the cosine similarity between image tokens and merge those exhibiting high similarity. Recently, Dynamic Tuning (DyT)~\cite{zhao2024dyt} keeps the pre-trained ViT parameters frozen, updating only the adapters and token dispatchers to enhance parameter efficiency and reduce redundant computation during inference. However, DyT cannot speed up the training procedure as it needs to use Gumbel Noise~\cite{herrmann2020channel} to determine what tokens to skip. Unlike DyT, our method can be combined with most prominent token sparsification strategies during training.

\subsection{Parameter-efficient Fine-tuning} 

With the trend of scaling up ViT~\cite{dosovitskiy2020vit,he2022mae,zhai2022vit-g,dehghani2023vit-22b} for enhanced performance and generalization, adapting entire models to downstream tasks becomes increasingly computationally prohibitive. To address this challenge, parameter-efficient fine-tuning (PEFT)~\cite{houlsby2019adapter,hu2021lora,jia2022vpt} has emerged as a strategic solution. PEFT methods update only a small subset of additional parameters while keeping the pre-trained model frozen, thereby mitigating risks of catastrophic forgetting and overfitting. Most PEFT methods designed for Transformers~\cite{vaswani2017transformer} fall into three categories: (1) \textbf{Partial Tuning}~\cite{kornblith2019linearprob,zaken2022bitfit}, which updates only a small subset of inherent parameters while freezing the majority; (2) \textbf{Prompt Tuning}~\cite{lester2021prompt,jia2022vpt,xin2024mmap}, which prepends learnable tokens (prompts) to the input and updates only these prompts during fine-tuning; (3) \textbf{Adapter Tuning}~\cite{houlsby2019adapter,chen2022adaptformer,xin2024vmt}, which inserts lightweight adapter modules and updates only their parameters during fine-tuning.


\begin{figure*}[t]
    \centering
    \includegraphics[width=\textwidth]{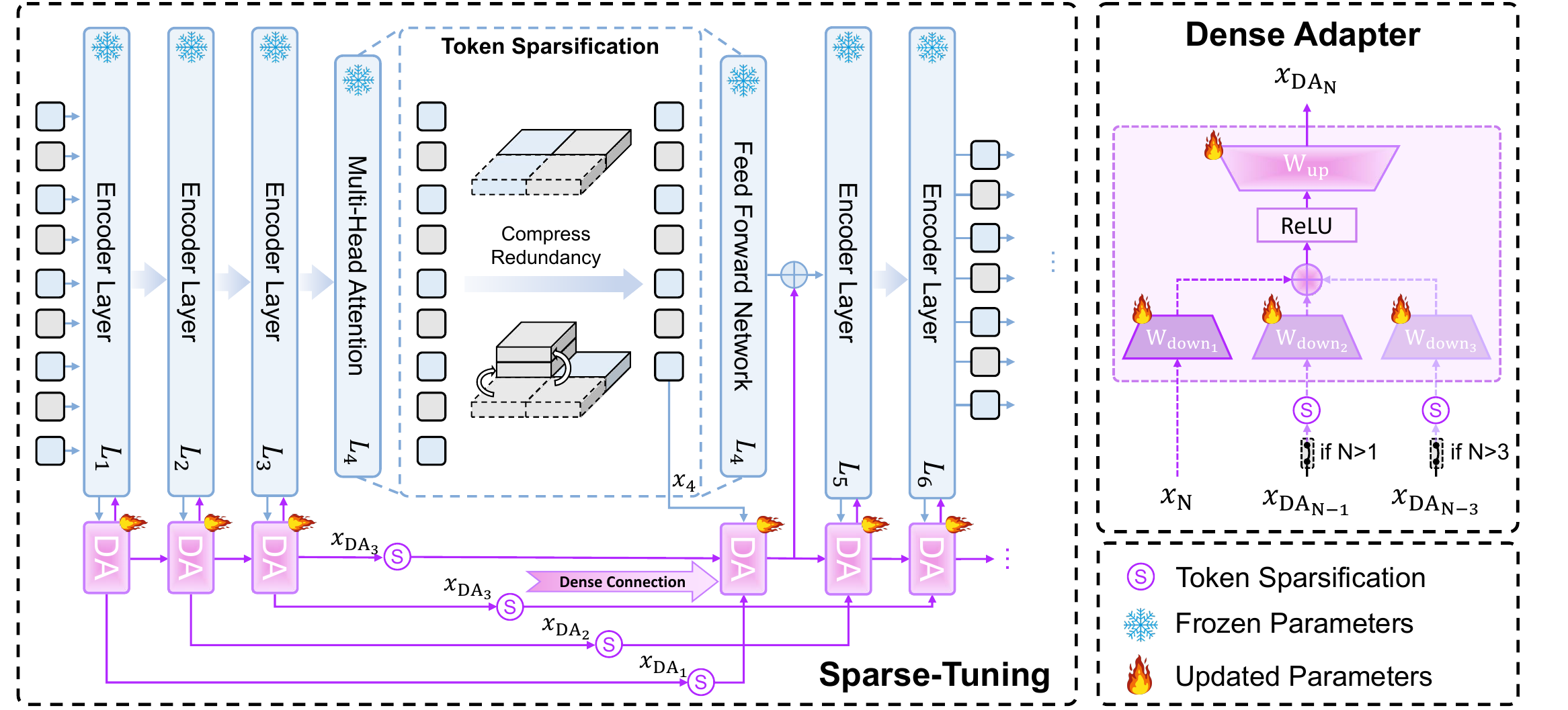}

    \caption{\textbf{Overall framework.} We freeze the pre-trained ViT-B/16 and update the proposed Dense Adapters (DAs) to efficiently fine-tune the pre-trained ViT. By selectively adapting tokens to focus on informative regions, Sparse-Tuning significantly reduces the computational cost of redundant tokens, thereby enhancing efficiency during both fine-tuning and inference stages. Token sparsification is the process of removing redundant tokens.}
    \vspace{-4mm}
\label{fig:overview}
\end{figure*}

While most PEFT methods improve parameter efficiency during fine-tuning, they often introduce new parameters that compromise inference efficiency. Reparameterization methods, such as LoRA \cite{hu2021lora} and FacT \cite{jie2023fact}, introduce learnable parameters that can be integrated into the original model during inference, thus maintaining the original model's inference efficiency. However, current PEFT methods do not account for maintaining performance under sparse token conditions, which is crucial for efficiently applying large-scale ViTs (\textit{e.g.}, ViT-L) to real-world applications. In this paper, we aim to improve both the fine-tuning and inference efficiency of pre-trained ViT.

%% file: sec/3_method.tex
\section{Methodology}
\label{Method}

Our method builds upon the Vision Transformer (ViT)~\cite{dosovitskiy2020vit} architecture. In the following, we first review the ViT architecture, adapter tuning techniques, and prominent token sparsification methods in Section~\ref{sub-sec:Preliminaries}. Subsequently, we investigate the integration of existing PEFT methods with token sparsification techniques in Section~\ref{method:Other PEFT+Token Compression}. Finally, we present our proposed Sparse-Tuning framework with carefully designed Dense Adapters (DAs) in Section~\ref{sub-sec:Sparse-Tuning}.

\subsection{Preliminaries}
\label{sub-sec:Preliminaries}

\paragraph{Vision Transformers} ViTs~\cite{dosovitskiy2020vit} basically consist of a patch embedding layer and a stack of transformer encoder layers. The patch embedding layer first splits and flattens an input image $\mathit{x} \in \mathbb{R}^{H \times W \times 3}$ into a sequence of patches $\mathit{x}_{p} \in \mathbb{R}^{N \times (P^2 \cdot C)}$, where $(H, W)$ denotes the size of the input image, $(P, P)$ represents the size of each image patch, $C$ is the number of channels, and $N = {H \cdot W}/{P^2}$ is the number of image tokens. The patches $\mathit{x}_{p}$, prepended with a learnable \texttt{[CLS]} token, are fed into a stack of transformer encoder layers, each of which includes a Multi-Head Attention (MHA) block and a Feed-Forward Network (FFN). In MHA, the tokens are linearly projected and packed into three vectors, namely $\bm{Q}$, $\bm{K}$, and $\bm{V}$. The self-attention operation can be written as:
\begin{equation}
\small
{\rm Attention}(\bm{Q}, \bm{K}, \bm{V}) = {\rm Softmax}(\frac{\bm{Q} \bm{K}^\top}{\sqrt{d}})\bm{V},
\end{equation}
${\rm Softmax}(\frac{\bm{Q} \bm{K}^\top}{\sqrt{d}})$ is the attention map, where $\bm{Q} \bm{K}^\top$ indicates the attention from the \texttt{[CLS]} token to all tokens and reflects the importance of each token. Subsequently, the output tokens are sent to a Layer Normalization (LayerNorm) \cite{ba2016layernorm} and a FFN, which consists of two fully-connected layers with a GELU activation function \cite{hendrycks2016gelu} in between. After processing the tokens by a stack of encoder layers, the \texttt{[CLS]} token is extracted and utilized for classification.

\paragraph{Adapter Tuning} Adapter Tuning~\cite{chen2022adaptformer} is a prevalent PEFT strategy that inserts lightweight modules in parallel with the FFN. Standard adapters comprise a down-projection layer $\mathbf{W}_{\text{down}}$, ReLU activation $\sigma(\cdot)$, and an up-projection layer $\mathbf{W}_{\text{up}}$. Given input feature $x$, the adapter function is expressed as:
\begin{equation}
\small
{\rm Adapter}(x) = x + s \cdot \sigma(x\mathbf{W}_{\text{down}})\mathbf{W}_{\text{up}},
\end{equation}
where $s$ denotes the scaling factor. In contrast to standard adapters, we propose Dense Adapters that aggregate multi-level features from different encoder layers, establishing dense connections across the network hierarchy.

\paragraph{Prominent Token Sparsification Methods}

To comprehensively evaluate our approach, we integrate various token sparsification techniques with Dense Adapters. We view token sparsification as a transformation:
\begin{equation}
\small
\tilde{X} = \mathcal{T}(X), \quad X = \{x_i\}_{i=1}^{N}, \quad \tilde{X} = \{\tilde{x}_m\}_{m=1}^{\tilde{N}}, \ \tilde{N} < N,
\end{equation}
where $X$ denotes the original token set, $\tilde{X}$ the sparsified tokens, and $\mathcal{T}$ is a token selection or aggregation operator. We instantiate $\mathcal{T}$ using pruning-based DynamicViT~\cite{rao2021dynamicvit} and merging-based EViT~\cite{liang2022evit} and ToMe~\cite{Bolya2023ToMe}.

Given the input token set $X = \{x_i\}_{i=1}^{N}$, \textbf{DynamicViT} learns a scalar importance score for each token via a lightweight prediction head $f_\theta$ and prunes tokens with low scores:
\begin{equation}
\small
s_i = f_\theta(x_i), \quad 
\mathcal{I}_{\text{keep}} = \operatorname{TopK}(s, \rho N),
\end{equation}
\begin{equation}
\small
\tilde{X} = \{x_i \mid i \in \mathcal{I}_{\text{keep}}\},
\end{equation}
where $s = [s_1,\dots,s_N]$ denotes the importance scores and $\rho \in (0,1]$ is the keep ratio.

\textbf{EViT} leverages the attention scores from the \texttt{[CLS]} token to select informative tokens. Let $a \in \mathbb{R}^{N}$ be the attention weights from \texttt{[CLS]} to all tokens, obtained from the row of the attention map corresponding to the \texttt{[CLS]} query,
\begin{equation}
\small
a = \operatorname{Softmax}\!\left(\frac{\bm{q}_{\text{cls}}\bm{K}^\top}{\sqrt{d}}\right),
\end{equation}
then EViT selects the top-$k$ informative tokens and fuses the remaining inattentive ones into a background token:
\begin{equation}
\small
\mathcal{I}_{\text{top}} = \operatorname{TopK}(a, k), \quad 
\bar{x} = \sum_{j \notin \mathcal{I}_{\text{top}}} \tilde{a}_j x_j,
\end{equation}
\begin{equation}
\small
\tilde{X} = \{x_i \mid i \in \mathcal{I}_{\text{top}}\} \cup \{\bar{x}\},
\end{equation}
where $\tilde{a}_j = \frac{a_j}{\sum_{l \notin \mathcal{I}_{\text{top}}} a_l}$ are normalized weights over the inattentive tokens.

In contrast, \textbf{ToMe} progressively merges similar tokens based on cosine similarity. It first computes a similarity matrix
\begin{equation}
\small
S_{ij} = \cos(x_i, x_j) = \frac{x_i x_j^\top}{\|x_i\|_2 \,\|x_j\|_2},
\end{equation}
then forms a set of high-similarity pairs $\mathcal{P}$ (\textit{e.g.}, via bipartite matching) and merges each pair $(i,j) \in \mathcal{P}$ into a single token:
\begin{equation}
\small
\tilde{x}_{(i,j)} = \frac{1}{2}(x_i + x_j),
\end{equation}
\begin{equation}
\small
\tilde{X} = \{\tilde{x}_{(i,j)} \mid (i,j) \in \mathcal{P}\} \cup \{x_k \mid k \notin \cup_{(i,j)\in\mathcal{P}}\{i,j\}\}.
\end{equation}

DynamicViT thus performs hard pruning of low-importance tokens, EViT selects top-$k$ tokens guided by \texttt{[CLS]} attention and aggregates the rest, while ToMe reduces token count by merging highly similar tokens rather than discarding them. We adopt EViT as our default TS scheme, primarily due to its high implementation efficiency.

\subsection{Combining PEFT with Token Sparsification}
\label{method:Other PEFT+Token Compression}

To investigate whether existing PEFT methods can effectively integrate with token sparsification techniques, we systematically explore various combinations. We select representative methods including merging-based ToMe~\cite{Bolya2023ToMe} and EViT~\cite{liang2022evit}, and pruning-based DynamicViT~\cite{rao2021dynamicvit}, combined with AdaptFormer~\cite{chen2022adaptformer} and LoRA~\cite{hu2021lora}. As shown in Table~\ref{other tc}, while these combinations improve efficiency compared to full fine-tuning, they suffer substantial performance degradation. This indicates that current PEFT methods inadequately preserve model performance under token sparsification. We attribute this to the inherent information loss from token sparsification, which disrupts the original attention distribution. Figure~\ref{attention} (c)(d) further confirms that conventional PEFT methods fail to restore the attention patterns. This observation raises a critical research question: \textit{How can we design fine-tuning strategies that effectively mitigate the information loss inherent in token sparsification while maintaining computational efficiency?}.

\input{Tables/code}

\subsection{Sparse-Tuning for Efficient ViT Adaption}
\label{sub-sec:Sparse-Tuning}

To answer this question, we first revisit the fundamental principle behind efficient vision models. Prior work~\cite{rao2021dynamicvit,wang2021dvt,liang2022evit} has established that ViT predictions predominantly rely on a subset of highly informative tokens, suggesting that careful token selection could maintain performance while improving efficiency. Building on this insight, DyT~\cite{zhao2024dyt} freezes pre-trained parameters and employs a token dispatcher within adapters to identify and discard uninformative tokens. However, this approach introduces \textbf{two critical problems} that prevent it from adequately addressing our research question: (1) \textit{Inefficient fine-tuning}—the token dispatcher requires gradients to backpropagate through all tokens, resulting in high GPU memory consumption and slow training; (2) \textit{Irreversible information loss}—directly removing tokens permanently discards potentially useful information, leading to performance degradation.

These limitations reveal that simply combining token selection with adapter tuning is insufficient. The key insight is that we need a mechanism to preserve and propagate information from discarded tokens rather than eliminating them entirely. Motivated by this analysis, we propose Sparse-Tuning with Dense Adapters, a novel approach that efficiently adapts pre-trained ViTs by selectively processing informative tokens while maintaining information flow through dense connections. As illustrated in Figure~\ref{fig:overview}, our framework comprises two components: 1) a frozen pre-trained ViT-B/16~\cite{dosovitskiy2020vit} with patch embedding and 12 encoder layers incorporating token sparsification, and 2) trainable Dense Adapters that preserve information across layers while adapting to downstream tasks.


\begin{figure}[t]
    \centering
    \includegraphics[width=0.47\textwidth]{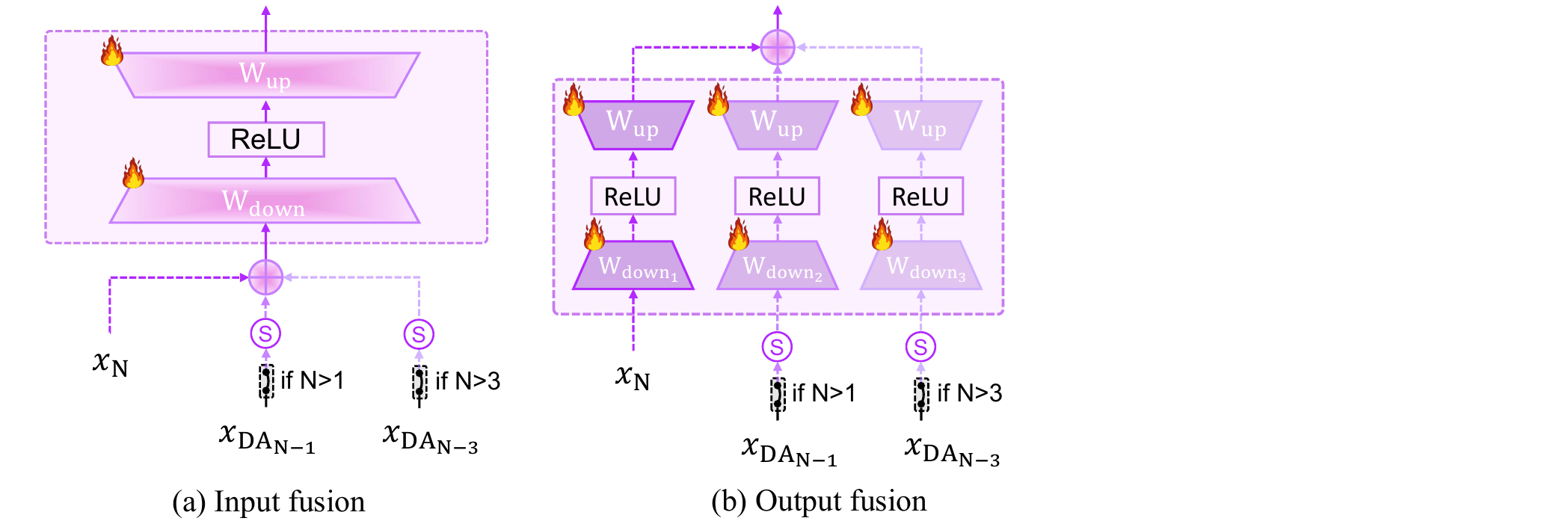}
    \caption{\textbf{Variants of Dense Adapters.} We present two variants that integrate multi-level features from different encoder layers at (a) the input stage and (b) the output stage.}
    \vspace{-4mm}
\label{fig:DA_variants}
\end{figure}

\paragraph{Token Sparsification with Dense Adapters} The core objective of Sparse-Tuning is to mitigate performance degradation in sparse token scenarios while achieving both computational efficiency and model effectiveness. To improve the fine-tuning efficiency of ViTs, we leverage existing token sparsification strategies as \textit{plug-and-play} modules, including the merging-based EViT~\cite{liang2022evit} and the pruning-based DynamicViT~\cite{rao2021dynamicvit}.
However, such sparsification inevitably incurs information loss (\textit{e.g.}, as observed in DyT~\cite{zhao2024dyt}).
To mitigate this degradation while efficiently adapting pre-trained ViTs to downstream tasks, we move beyond conventional adapter-tuning paradigms and propose a more effective adaptation mechanism.

Traditional adapter-tuning methods for ViT~\cite{houlsby2019adapter,chen2022adaptformer,jiang2024res-tuning} follow ResNet's~\cite{he2016resnet} \textit{residual connection} paradigm, which restricts connections within individual layers and limits inter-layer feature propagation during fine-tuning. Moreover, when deeper layers retain only semantic-relevant but spatially coarse tokens, this design fails to recover the fine-grained details lost through token sparsification.

Inspired by DenseNet~\cite{huang2017densenet}, we introduce Dense Adapters (DAs) that establish \textit{dense connections} from shallow to deep layers, enabling the propagation of local detailed features. As illustrated in Figure~\ref{fig:overview} (right), by aggregating multi-level features from different layers, DAs transform the preserved tokens into enriched representations with high-resolution characteristics.

Specifically, each DA dynamically incorporates one to three down-projection layers ($\mathbf{W}_{\text{down}_1}$, $\mathbf{W}_{\text{down}_2}$, $\mathbf{W}_{\text{down}_3}$), ReLU activation, and an up-projection layer $\mathbf{W}_{\text{up}}$. For the N-th DA (denoted as $\text{DA}_\text{N}$), the output is formulated as:
\begin{equation}
\small
x_{\text{DA}_\text{N}} = \sigma(x_\text{down})\mathbf{W}_{\text{up}},
\end{equation}
where $x_\text{down}$ is calculated by:
\begin{equation}
\small
x_\text{down} = 
\begin{cases} 
x_\text{N}\mathbf{W}_{\text{down}_1}, \text{if N} = 1,   \\     
x_\text{N}\mathbf{W}_{\text{down}_1} + x_{\text{DA}_\text{N-1}}\mathbf{W}_{\text{down}_2}, \text{if N} \in (1, 3], \\
x_\text{N}\mathbf{W}_{\text{down}_1} + x_{\text{DA}_\text{N-1}}\mathbf{W}_{\text{down}_2} + x_{\text{DA}_\text{N-3}}\mathbf{W}_{\text{down}_3}, \text{if N} > 3, \\
\end{cases}
\end{equation}
where $x_{\text{DA}_\text{N}}$ and $x_\text{N}$ denote the outputs of $\text{DA}_\text{N}$ and the MHA at layer N, respectively. Note that for $3 < \text{N} \leq 12$, features $x_{\text{DA}_{\text{N}-1}}$ and $x_{\text{DA}_{\text{N}-3}}$ undergo token sparsification to maintain consistent dimensions across all inputs. Through dense connections, our DAs effectively compensate for information loss while enhancing the model's representational capacity.

\input{Tables/complete_datasets}

\paragraph{Dense Adapter Variants} To explore optimal multi-level feature fusion strategies, we design two architectural variants, as illustrated in Figure~\ref{fig:DA_variants}. These variants differ primarily in their feature integration approach.

Variant (a) performs early fusion by combining multi-level features before adapter processing:
\begin{equation}
\small
x_{\text{fusion}} = x_\text{N} + x_{\text{DA}_{\text{N}-1}} + x_{\text{DA}_{\text{N}-3}},
\end{equation}
\begin{equation}
\small
\begin{aligned}
x_\text{down} &= x_{\text{fusion}}\mathbf{W}_{\text{down}_1},\\
x_{\text{DA}_\text{N}} &= \sigma(x_\text{down})\mathbf{W}_{\text{up}}.
\end{aligned}
\end{equation}

Variant (b) processes features independently through parallel pathways before late fusion:
\begin{equation}
\small
\begin{aligned}
x_1 &= \sigma(x_\text{N}\mathbf{W}_{\text{down}_1})\mathbf{W}_{\text{up}_1},\\
x_2 &= \sigma(x_{\text{DA}_{\text{N}-1}}\mathbf{W}_{\text{down}_2})\mathbf{W}_{\text{up}_2},\\
x_3 &= \sigma(x_{\text{DA}_{\text{N}-3}}\mathbf{W}_{\text{down}_3})\mathbf{W}_{\text{up}_3},
\end{aligned}
\end{equation}
\begin{equation}
\small
x_{\text{DA}_\text{N}} = x_1 + x_2 + x_3.
\end{equation}

We present the algorithm of Sparse-Tuning in Algorithm ~\ref{alg:pseudo} to help better understand the whole process.

%% file: Tables/code.tex
\begin{algorithm}[t!]
\caption{Sparse-Tuning for a ViT Encoder}
\label{alg:pseudo}
\begin{algorithmic}[1]
\Require Vision tokens $\mathbf{x} \in \mathbb{R}^{N \times C}$, token keeping rate $k$, number of heads $H$, layer index $n$
\Ensure Output tokens $\mathbf{x}$ after sparse-tuning
\State \textbf{Step 1: Parameter Freezing}
\For{each parameter $p$ in model}
    \If{``adapter'' $\in$ $p$.name \textbf{or} ``head'' $\in$ $p$.name}
        \State $p$.requires\_grad $\leftarrow$ \textbf{True}
    \Else
        \State $p$.requires\_grad $\leftarrow$ \textbf{False}
    \EndIf
\EndFor
\State \textbf{Step 2: Multi-Head Attention Computation}
\State Initialize avg\_cls\_attn $\leftarrow$ zeros($N-1$)
\For{head $i = 1$ to $H$}
    \State $\mathbf{Q}_i \leftarrow \mathrm{fc}_q^{(i)}(\mathbf{x})$, $\mathbf{K}_i \leftarrow \mathrm{fc}_k^{(i)}(\mathbf{x})$, $\mathbf{V}_i \leftarrow \mathrm{fc}_v^{(i)}(\mathbf{x})$
    \State $\mathbf{A}_i \leftarrow \mathrm{softmax}\left(\frac{\mathbf{Q}_i \mathbf{K}_i^\top}{\sqrt{C/H}}\right)$
    \State $\mathbf{H}_i \leftarrow \mathbf{A}_i \mathbf{V}_i$
    \State avg\_cls\_attn $\leftarrow$ avg\_cls\_attn $+$ $\mathbf{A}_i[0, 1:]$
\EndFor
\State $\mathbf{x} \leftarrow \mathrm{concat}(\mathbf{H}_1, \ldots, \mathbf{H}_H)$
\State $\mathbf{x} \leftarrow \mathrm{proj}(\mathbf{x}) + \mathbf{x}_{\text{residual}}$
\State \textbf{Step 3: Token Sparsification}
\State avg\_cls\_attn $\leftarrow$ avg\_cls\_attn / $H$
\State indices $\leftarrow \mathrm{argsort}(\text{avg\_cls\_attn}, \text{descending})$
\State $K \leftarrow \lceil k \cdot (N-1) \rceil$
\State topk\_idx $\leftarrow$ indices$[:K]$, non\_topk\_idx $\leftarrow$ indices$[K:]$
\State $\mathbf{x}_{\text{cls}} \leftarrow \mathbf{x}[0]$, $\mathbf{x}_{\text{tokens}} \leftarrow \mathbf{x}[1:]$
\State $\mathbf{x}_{\text{attentive}} \leftarrow \mathbf{x}_{\text{tokens}}[\text{topk\_idx}]$
\State $\mathbf{x}_{\text{inattentive}} \leftarrow \mathbf{x}_{\text{tokens}}[\text{non\_topk\_idx}]$
\State $\mathbf{x}_{\text{fused}} \leftarrow \mathrm{weighted\_mean}(\mathbf{x}_{\text{inattentive}}, \text{avg\_cls\_attn}[K:])$
\State $\mathbf{x} \leftarrow \mathrm{concat}([\mathbf{x}_{\text{cls}}, \mathbf{x}_{\text{attentive}}, \mathbf{x}_{\text{fused}}])$
\State \textbf{Step 4: Dense Adapter and Feed-Forward}
\State $\mathbf{x}_{\text{adapter}} \leftarrow \mathrm{dense\_adapter}(\mathbf{x}, \mathbf{x}_{\text{prev}}, \mathbf{x}_{n-3})$
\State $\mathbf{x}_{\text{res}} \leftarrow \mathbf{x}$
\State $\mathbf{x} \leftarrow \mathrm{norm}(\mathbf{x})$
\State $\mathbf{x} \leftarrow \mathrm{ffn}(\mathbf{x})$
\State $\mathbf{x} \leftarrow \mathbf{x} + \mathbf{x}_{\text{res}} + \mathbf{x}_{\text{adapter}}$
\State \Return $\mathbf{x}$
\end{algorithmic}
\end{algorithm}

%% file: Tables/complete_datasets.tex
\begin{table*}[h]
    \caption{\textbf{Results on complete image and video datasets.} Avg.: the mean value from results across various image and video datasets. GFLOPs are evaluated on CIFAR-100 and K400. DyT$\dag$ $N=4$ represents DyT with four experts.}
    \label{tab:adaptformer_benchmark}
\addtolength{\tabcolsep}{-2pt}
    \centering
    \small

    \begin{tabular}{l | c  | c c c c c | c c c c}
    \toprule
    \multirow{2}{*}{Method} & Params.  $\downarrow$  & \multicolumn{5}{c|}{\textbf{Image Datasets}} & \multicolumn{4}{c}{\textbf{Video Datasets}} \\ 
    & (M) & GFLOPs $\downarrow$ & CIFAR-100 & SVHN & Food-101 & Avg. & GFLOPs $\downarrow$ & K400 & SSv2 & Avg. \\
    \midrule

    \multicolumn{11}{c}{\textit{Traditional fine-tuning}} \\ 
    \midrule
    Full fine-tuning & 85.80 & 17.58 & 90.91 & 97.29 & 90.69 & 92.69 & 142.53 & 75.48 & 75.22 & 60.35 \\
    Linear & \textbf{0} & 17.58 & 85.87 & 56.29 & 88.07 & 76.74 & 142.53 & 69.04 & 27.64 & 48.34  \\
    \midrule
    \multicolumn{11}{c}{\textit{Parameter-efficient fine-tuning}} \\ 
    \midrule
    Adapter \cite{houlsby2019adapter} [ICML'19] & 1.19 & 17.81 & 91.76 & 96.88 & 89.91 & 92.76 & 144.39 & 74.72 & 44.58 & 59.75 \\
    
    AdaptFormer \cite{chen2022adaptformer} [NeurlPS'22] & 1.19 & 17.81 & \underline{92.03} & 97.23 & \textbf{90.84} & \underline{93.36} & 144.39 & \underline{75.53} & 45.36 & 60.45 \\
    LoRA \cite{hu2021lora}[ICLR'22]  & 1.19 & 17.58 & 91.42 & \underline{97.36} & 90.48 & 93.08 & 142.53 & 75.48 & 45.62 & 60.55 \\
    VPT \cite{jia2022vpt} [ECCV'22] & \underline{0.07} & 18.32 & 91.64 & 95.72 & 90.41 & 92.59 & 148.44 & 73.46 & 38.17 & 55.82 \\
    DyT \cite{zhao2024dyt} [NeurlPS'24]& 1.19 & \underline{12.21} & 91.37 & 97.08 & 90.32 & 92.92 & 108.31 & 74.39 & 45.34 & 59.87 \\
    DyT$\dag$ $N=4$ \cite{zhao2024dyt}[NeurlPS'24] & 4.80 & 12.29 & 91.01 & 96.90 & 89.77 & 92.56 & \underline{105.45} & 75.00 & \underline{46.56} & \underline{60.78} \\
    \midrule

    \multicolumn{11}{c}{\textit{The proposed Sparse-Tuning}} \\ 
    \midrule
    \rowcolor{tabhighlight}
    Sparse-Tuning &  1.10  & \textbf{11.70} &  \textbf{92.31}  &  \textbf{97.47}  &  \underline{90.72}  &  \textbf{93.50}  &  \textbf{99.80}  &  \textbf{75.55}  &  \textbf{46.67} & \textbf{61.11} \\
    
    \bottomrule
    
    \end{tabular}

\end{table*}

%% file: sec/4_exp.tex
\section{Experiments}
\label{Experiments}

\input{Tables/vtab1k}

\input{Tables/CIFAR_100}
\input{Tables/sparse_comparisons}

\subsection{Experimental Setup}
\label{sub-sec:Setup}

\paragraph{Datasets} We comprehensively evaluate our Sparse-Tuning against existing methods on the widely-adopted PEFT benchmark VTAB-1K~\cite{zhai2019vtab}, which specifically assesses adaptation performance under limited training data constraints. VTAB-1K presents a challenging scenario in which each downstream classification task contains only 1,000 training samples, testing the data efficiency of fine-tuning methods. To provide a more thorough evaluation beyond this low-data regime, we follow~\cite{chen2022adaptformer,zhao2024dyt} and conduct extensive experiments on three complete image classification datasets: CIFAR-100~\cite{krizhevsky2009cifar-100} with 100 fine-grained categories, SVHN~\cite{goodfellow2013svhn} for digit recognition in natural scenes, and Food-101~\cite{bossard2014food-100} containing 101 food categories. Additionally, we evaluate on two challenging video action recognition datasets, namely Kinetics-400 (K400)~\cite{carreira2017k400} with 400 human action classes and Something-Something V2 (SSv2)~\cite{goyal2017ssv2} focusing on temporal reasoning, to comprehensively assess both adaptation performance and computational efficiency across diverse visual domains.

\paragraph{Implementation Details} We adopt ViT-Base (ViT-B/16)~\cite{dosovitskiy2020vit} as our backbone architecture, initialized with weights pre-trained on the ImageNet-21K dataset~\cite{deng2009imagenet} containing 14 million images across 21,841 classes. For our Dense Adapter configuration, the bottleneck dimension $d$ is set to 32 by default for full datasets, while we reduce it to 8 specifically for VTAB-1K experiments to match the settings of prior work~\cite{chen2022adaptformer,zhao2024dyt} and accommodate the limited training data. The scaling factor $s$ is consistently set to 1.0 across all experiments. Regarding TS, we set the retention rate $r$ of semantically-relevant tokens to 0.7 (\textit{i.e.}, preserving 70\% of tokens) by default, unless otherwise specified in ablation studies. All experiments are conducted on NVIDIA A800 GPUs with mixed precision training to optimize memory usage and computational efficiency. We use the AdamW optimizer~\cite{adamw} with a cosine learning rate schedule and apply standard data augmentation techniques following established protocols for each dataset.

\subsection{Main Results}
\label{sub-sec:Main results}

\paragraph{Comparisons on Complete Datasets} We conduct experiments on comprehensive image and video datasets to evaluate the adaptation performance with abundant training data. The results on complete image and video datasets are shown in Table \ref{tab:adaptformer_benchmark}, from which we find that: \textbf{(1)} Sparse-Tuning outperforms all baseline methods on both image and video datasets, demonstrating its strong transferability on complete datasets. Taking DyT (N = 4) in image datasets as an example, Sparse-Tuning outperforms it by 0.94 in average performance. Specifically, it surpasses DyT (N = 4) by 1.3, 0.57, and 0.95 on the CIFAR-100, SVHN, and Food-101 datasets respectively.\textbf{(2)} Sparse-Tuning demonstrates exceptional inference efficiency on both image and video datasets. Particularly on video datasets, Sparse-Tuning reduces the computational complexity of the original ViT-B by around 30\%, highlighting its strong efficiency in video applications. With only 1.10M updated parameters, Sparse-Tuning achieves superior performance in image and video recognition, while significantly improving efficiency.

\paragraph{Comparisons on VTAB-1K} The comparison results with state-of-the-art PEFT methods on VTAB-1K \cite{zhai2019vtab} are presented in Table \ref{taba:vtab1k}, from which we can observe that: \textbf{(1)} Sparse-Tuning outperforms all SOTA PEFT methods. Sparse-Tuning achieves better performance in terms of average accuracy across the entire dataset compared to the previous best model DyT \cite{zhao2024dyt}. \textbf{(2)} Sparse-Tuning largely improves inference efficiency. With only 11.65 GFLOPs, about 66\% of the computational cost of the original ViT-B, Sparse-Tuning with keeping rate $r=0.7$ has outperformed all state-of-the-art methods in terms of both performance and inference efficiency. \textbf{(3)} Sparse-Tuning continues to exhibit better performance when the keeping rate $r$ increases.

\input{Tables/ablation}

\input{Tables/feature_input}
\input{Tables/feture_fusion}

\input{Tables/token_drop}

\paragraph{Detailed Performance and Efficiency Analysis} In Table~\ref{tab:CIFAR-100}, we provide a comprehensive analysis of the performance–efficiency trade-offs across various PEFT methods. Specifically, we report the number of updated parameters during fine-tuning, GPU memory consumption during both fine-tuning and inference, wall-clock time for fine-tuning and inference, computational cost in GFLOPs, and final accuracy on the complete CIFAR-100~\cite{krizhevsky2009cifar-100}. Compared to mainstream PEFT baselines, our Sparse-Tuning achieves state-of-the-art accuracy while significantly improving efficiency by reducing memory footprint, computational cost, and runtime during both fine-tuning and inference.

\paragraph{Comparison and Compatibility with Methods for Efficient ViT} We systematically integrate multiple TS methods with mainstream PEFT approaches and analyze their performance on the VTAB-1K benchmark. As shown in Table~\ref{other tc}, Sparse-Tuning exhibits superior compatibility with TS techniques compared to AdaptFormer, Adapter, and LoRA. Notably, when integrated with ToMe, Sparse-Tuning achieves 75.66\% accuracy while maintaining higher efficiency than other methods, demonstrating an optimal trade-off between computational efficiency and model performance.

\subsection{Ablation Studies}
\label{sub-sec:Ablation}

In this subsection, we first analyze the effectiveness of TS and Dense Adapters. Next, we provide an in-depth examination of the feature inputs and fusion methods of Dense Adapters. Subsequently, we explore the impact of different strategies for processing semantic-irrelevant tokens. Finally, we verify the effectiveness of Sparse-Tuning when scaling up the ViT.

\paragraph{Components Effectiveness} In Table \ref{Table:TS and DA}, we report the performance of using different components of Sparse-Tuning to investigate the effectiveness of TS and Dense Adapters. We can observe the following: \textbf{(1)} TS can reduce the computational complexity, but it leads to a significant performance degradation, resulting in a 7\% decrease in average accuracy (Table \ref{Table:TS and DA} (a,b)). \textbf{(2)} Dense Adapters can significantly improve the performance across three datasets (Table \ref{Table:TS and DA} (a,c)), which demonstrates their effectiveness in ViT adaption. \textbf{(3)} In Table \ref{Table:TS and DA} (c,d), we observe that Sparse-Tuning (d) sacrifices an average of 0.48\% accuracy compared to using only Dense Adapters (c), indicating that although TS significantly reduces ViT computational costs, it may incur a minor performance drop. Nonetheless, considering the performance and efficiency trade-off, Sparse-Tuning achieves the best balance.

\input{Tables/sparse_layer}
\input{Tables/neck}
\input{Tables/boundary}

\begin{figure*}[t]
    \centering
    \includegraphics[width=\textwidth]{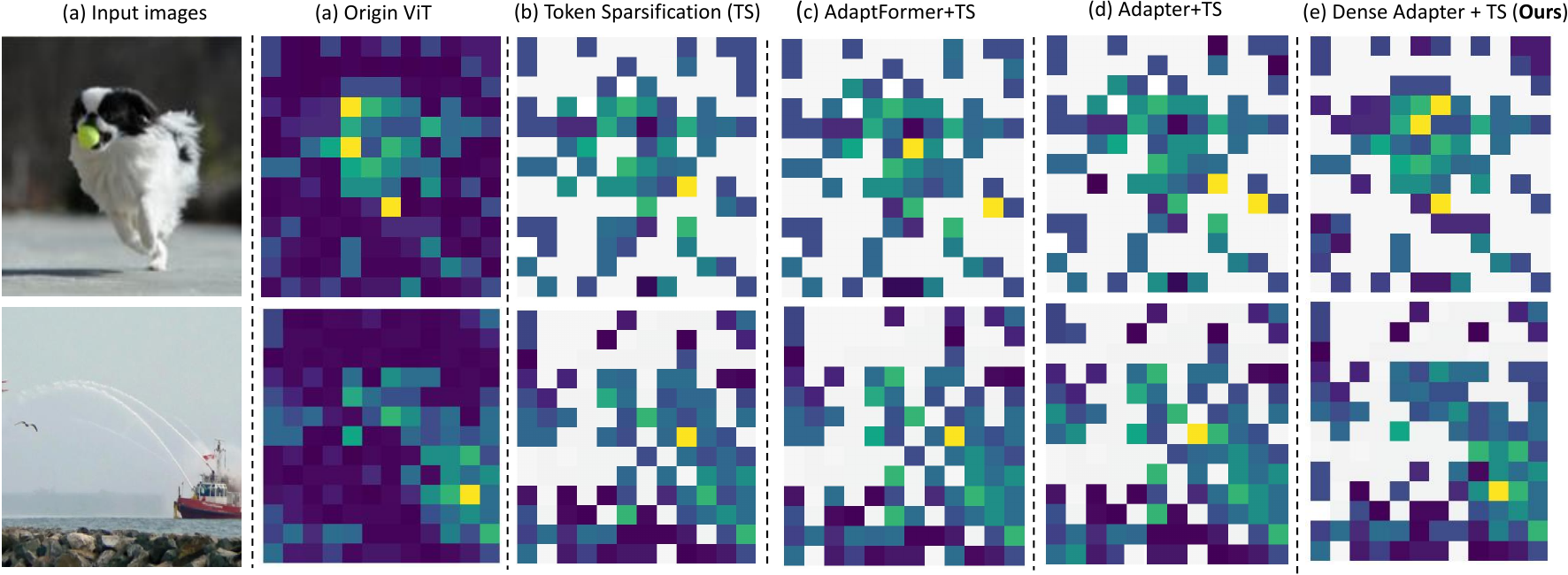}

    \vspace{-2mm}
    \caption{Comparison of attention maps between our method and other PEFT+TS combinations.} 
    \vspace{-4mm}
\label{attention}
\end{figure*}

\input{Tables/interval}

\paragraph{Effects of Different Feature Inputs} To investigate the effectiveness of dense connections, we compare different inputs to Dense Adapters. As shown in Table \ref{Table:feature input}, when feeding multiple features from different encoder layers into the Dense Adapters, the performance increases. This suggests that Dense Adapters effectively facilitate dense interactions between the lower and higher layers of the ViT to enhance the representational capability, thereby improving performance compared to adapter-tuning (Table \ref{Table:feature input} (a)). Notably, while Sparse-Tuning introduces more feature interactions requiring computation, the GFLOPs are still reduced compared to adapter-tuning, suggesting that TS also reduces the computation cost in Dense Adapters.

\paragraph{Effects of Different Feature Fusion Methods} Since Dense Adapters take multiple features as inputs, we consider three variants of Dense Adapters that can fuse these multi-level features in Figure \ref{fig:DA_variants}. We report the performance of different feature fusion methods in Table \ref{Table:feature fusion}. Fusing the multi-level features before feeding them into the Dense Adapters (Figure \ref{fig:DA_variants} (a)) requires fewer trainable parameters but deteriorates performance. This occurs because this fusion method leads to information loss; features from different layers may contain complementary information, and simple addition may not effectively integrate this information. Fusing the features after feeding them into the Dense Adapters (Figure \ref{fig:DA_variants} (b)) also deteriorates performance. This is because multi-level features are mapped into different spaces, and directly fusing them may obscure important information, thereby reducing classification performance. Dense Adapters first project multi-level features into the same space, then fuse them, and finally up-project the fused features back into their original shape (Figure \ref{fig:overview}). This ensures that the dense interaction process occurs within the same feature space, which leads to better performance.

\paragraph{Effects of Different Irrelevant Token Processing Strategies} In Table \ref{Table:Token drop}, we evaluate the impact of three different strategies for handling semantic-irrelevant tokens in Sparse-Tuning. Clearly, Sparse-Tuning with irrelevant token "Merge" and "Argmax" both outperform the "Drop" strategy, indicating that semantic-irrelevant tokens do contribute to classification. Additionally, both of "Merge" and "Argmax" strategies introduce no additional computational overhead compared with directly dropping the semantic-irrelevant tokens. Notably, merging the irrelevant tokens into a single representative token may aggregate the information from all irrelevant tokens, thereby achieving the best performance.

\paragraph{Different Positions of TS} Since TS occurs across different layers in the ViT, we investigate its optimal positioning for performance-efficiency trade-off. As shown in Table \ref{Table:sparsification position}, earlier TS reduces computational cost by processing fewer full-token layers. However, early-stage sparsification makes subsequent attention computations \textit{less reliable}, degrading performance (Table \ref{Table:sparsification position} (a, b)). Later-stage sparsification (d, e) performs better but still sub-optimally due to lost local features. We find that TS in the 4th, 7th, and 10th encoder layers achieves the best results, suggesting that middle-layer dense interactions better balance local and global features. Therefore, we select these layers for optimal performance-efficiency trade-off.

\paragraph{Different Bottleneck Dimensions of DA} We explore the impact of the bottleneck dimension $d$ of DA to achieve the best trade-off between performance, updated parameters, and computational cost. As reported in Table \ref{Table:neck}, a higher bottleneck dimension $d$ introduces more parameters and higher GFLOPs. However, with a smaller $d$, the down-projection may lose significant information about the original features, leading to performance degradation. We observe that performance peaks at a bottleneck dimension of 32 and declines thereafter. Therefore, considering the trade-off between trainable parameters, GFLOPs, and performance, we set $d=32$.

\input{Tables/ViT_L}

\input{Tables/segmentation}

\paragraph{Different Starting Layers of TS} We explore different starting layers of TS to explore the optimal performance. From Table \ref{Table:sparsification boundary} (a,b), applying TS too early in shallow layers reduces accuracy, as ViT struggles to identify important tokens at these early stages due to unreliable attention maps. Similarly, from Table \ref{Table:sparsification boundary} (b,c), starting TS too late in deeper layers also hurts accuracy, likely due to the loss of key token interactions. Therefore, we select to conduct TS in the 4-th encoder layers.

\paragraph{Different Interval of Layers for TS} We further explore the impact of different intervals between encoder layers when implementing TS, starting from the 4th encoder layer. From Table \ref{Table:sparsification interval}, we observe that shorter intervals (\textit{e.g.}, interval = 2) may lead to more frequent losses of visual information, thereby impairing performance. Conversely, longer intervals (\textit{e.g.}, interval = 4) may allow irrelevant tokens to persist for too long, leading to an accumulation of noise and less effective token reduction, which ultimately compromises feature representation and reduces accuracy. Based on these observations, we select an interval of 3 encoder layers in practical applications, achieving a balance between minimizing information loss and reducing irrelevant token retention.


\subsection{Extended Capabilities of Sparse-Tuning}
\label{sub-sec:Extended Capabilities}

We further validate the effectiveness of Sparse-Tuning under more scenarios, including larger backbone networks and different vision tasks like segmentation.

\paragraph{Scaling to Larger ViT} We evaluate Sparse-Tuning's scalability by applying it to ViT-L \cite{dosovitskiy2020vit}. As shown in Table \ref{Table:vit-l}, Sparse-Tuning significantly reduces tunable parameters by \textbf{99.03\%} and decreases GFLOPs by \textbf{12.69-50.28\%} compared to full fine-tuning, while achieving superior performance. Moreover, Sparse-Tuning demonstrates better efficiency-performance trade-off than DyT \cite{zhao2024dyt}, confirming its effectiveness on larger pre-trained models.


\paragraph{Generalization to Segmentation Tasks} To demonstrate Sparse-Tuning's versatility across different vision tasks, we evaluate it on open-vocabulary segmentation task using the state-of-the-art model ProxyCLIP~\cite{lan2024ProxyClip}. As shown in Table~\ref{tab:seg}, Sparse-Tuning outperforms both full fine-tuning and DyT, with notable improvements of \textbf{1.6\%} and \textbf{2.2\%} over DyT on Object and VOC20 datasets, respectively.

\subsection{Qualitative Results}
\label{subsec:Qualitative Analysis}

Figure~\ref{attention} visualizes the attention scores between the patch tokens in the image and the \texttt{[CLS]} token, where brighter regions indicate higher attention scores, reflecting the greater contribution of a patch token to image classification. Through the attention map comparison in Figure~\ref{attention}, we investigate how different PEFT strategies influence the attention distribution when TS is introduced into the ViT architecture. As shown in Figure~\ref{attention} (b), token sparsification significantly removes background tokens, but the high-response regions still differ from those in the original ViT. When applying conventional PEFT variants, such as AdaptFormer and Adapter (Figures~\ref{attention} (c) and (d)), the high-response areas continue to differ from those in the original ViT’s attention. In contrast, Figure~\ref{attention} (e) shows that the Dense Adapter in our sparse-tuning approach effectively compensates for the information loss caused by TS, restoring the attention distribution closer to the original ViT.

%% file: Tables/vtab1k.tex
\begin{table*}[t]
\caption{\textbf{Comparison to state-of-the-art PEFT methods on VTAB-1K.} All Mean: the average accuracy of all datasets. Params.: the number of learnable parameters excluding the final classification layer. GFLOPs: the average GFLOPs across all datasets. $r$ denotes the keeping rate of attentive (activated) tokens. We highlight the \textbf{best} and the \underline{second-best} results.}
\label{taba:vtab1k}
\centering
\scriptsize
\setlength{\tabcolsep}{3pt}

\begin{tabular}{l | ccccccc | cccc | cccccccc | ccc}
\toprule
& \multicolumn{7}{c|}{\textbf{Natural}} & \multicolumn{4}{c|}{\textbf{Specialized}} & \multicolumn{8}{c|}{\textbf{Structured}} \\
  & \rotatebox{90}{\raisebox{0.5pt}{\tikz\fill[natural] (0,0) circle (.5ex);} CIFAR-100}
 & \rotatebox{90}{\raisebox{0.5pt}{\tikz\fill[natural] (0,0) circle (.5ex);} Caltech101}
 & \rotatebox{90}{\raisebox{0.5pt}{\tikz\fill[natural] (0,0) circle (.5ex);} DTD}
 & \rotatebox{90}{\raisebox{0.5pt}{\tikz\fill[natural] (0,0) circle (.5ex);} Flowers102}
 & \rotatebox{90}{\raisebox{0.5pt}{\tikz\fill[natural] (0,0) circle (.5ex);} Pets}
 & \rotatebox{90}{\raisebox{0.5pt}{\tikz\fill[natural] (0,0) circle (.5ex);} SVHN}
 & \rotatebox{90}{\raisebox{0.5pt}{\tikz\fill[natural] (0,0) circle (.5ex);} Sun397}
 & \rotatebox{90}{\raisebox{0.5pt}{\tikz\fill[specialized] (0,0) circle (.5ex);} Camelyon}
 & \rotatebox{90}{\raisebox{0.5pt}{\tikz\fill[specialized] (0,0) circle (.5ex);} EuroSAT}
 & \rotatebox{90}{\raisebox{0.5pt}{\tikz\fill[specialized] (0,0) circle (.5ex);} Resisc45}
 & \rotatebox{90}{\raisebox{0.5pt}{\tikz\fill[specialized] (0,0) circle (.5ex);} Retinopathy}
 & \rotatebox{90}{\raisebox{0.5pt}{\tikz\fill[structured] (0,0) circle (.5ex);} Clevr-Count}
 & \rotatebox{90}{\raisebox{0.5pt}{\tikz\fill[structured] (0,0) circle (.5ex);} Clevr-Dist}
 & \rotatebox{90}{\raisebox{0.5pt}{\tikz\fill[structured] (0,0) circle (.5ex);} DMLab}
 & \rotatebox{90}{\raisebox{0.5pt}{\tikz\fill[structured] (0,0) circle (.5ex);} KITTI-Dist}
 & \rotatebox{90}{\raisebox{0.5pt}{\tikz\fill[structured] (0,0) circle (.5ex);} dSpr-Loc}
 & \rotatebox{90}{\raisebox{0.5pt}{\tikz\fill[structured] (0,0) circle (.5ex);} dSpr-Ori}
 & \rotatebox{90}{\raisebox{0.5pt}{\tikz\fill[structured] (0,0) circle (.5ex);} sNORB-Azim}
 & \rotatebox{90}{\raisebox{0.5pt}{\tikz\fill[structured] (0,0) circle (.5ex);} sNORB-Elev}
 & \rotatebox{90}{\raisebox{0.5pt}{\tikz\fill[vtabmean] (0,0) circle (.5ex);} All Mean}  
 & \rotatebox{90}{\raisebox{0.5pt}{\tikz\fill[vtabparam] (0,0) circle (.5ex);} Params. (M) $\downarrow$} 
 & \rotatebox{90}{\raisebox{0.5pt}{\tikz\fill[vtabparam] (0,0) circle (.5ex);} GFLOPs $\downarrow$} \\
\midrule

\multicolumn{23}{c}{\emph{Traditional fine-tuning}}\\
\midrule
Full fine-tunig & 68.9 & 87.7 & 64.3 & 97.2 & 86.9 & 87.4 & 38.8 & 79.7 & 95.7 & 84.2 & 73.9 & 56.3 & 58.6 & 41.7 & 65.5 & 57.5 & 46.7 & 25.7 & 29.1 & 68.96 & 85.84 & 17.58 \\
Linear & 63.4 & 85.0 & 63.2 & 97.0 & 86.3 & 36.6 & 51.0 & 78.5 & 87.5 & 68.6 & 74.0 & 34.3 & 30.6 & 33.2 & 55.4 & 12.5 & 20.0 & 9.6 & 19.2 & 57.64 & \textbf{0} & 17.58 \\
\midrule

\multicolumn{23}{c}{\emph{Parameter-efficient fine-tuning}}\\
\midrule
Adapter~\cite{houlsby2019adapter} [ICML'19] & 69.2 & 90.1 & 68.0 & 98.8 & 89.9 & 82.8 & 54.3 & 84.0 & 94.9 & 81.9 & 75.5 & 80.9 & 65.3 & 48.6 & 78.3 & 74.8 & 48.5 & 29.9 & 41.6 & 71.44 & 0.16 & 17.61 \\
BitFit~\cite{zaken2022bitfit} [ACL'22] & 72.8 & 87.0 & 59.2 & 97.5 & 85.3 & 59.9 & 51.4 & 78.7 & 91.6 & 72.9 & 69.8 & 61.5 & 55.6 & 32.4 & 55.9 & 66.6 & 40.0 & 15.7 & 25.1  & 62.05 &  0.10 & 17.58 \\
LoRA~\cite{hu2021lora} [ICLR'22] & 67.1 & 91.4 & 69.4 & 98.8 & 90.4 & 85.3 & 54.0 & 84.9 & 95.3 & 84.4 & 73.6 & 82.9 & \textbf{69.2} & 49.8 & 78.5 & 75.7 & 47.1 & 31.0 & 44.0 & 72.25 & 0.29 & 17.58 \\
VPT~\cite{jia2022vpt} [ECCV'22] & \textbf{78.8} & 90.8 & 65.8 & 98.0 & 88.3 & 78.1 & 49.6 & 81.8 & 96.1 & 83.4 & 68.4 & 68.5 & 60.0 & 46.5 & 72.8 & 73.6 & 47.9 & 32.9 & 37.8 & 69.43  & 0.53 & 18.30 \\
SSF~\cite{lian2022ssf} [NeurIPS'22] & 69.0 & 92.6 & \textbf{75.1} & \underline{99.4} & 91.8 & 90.2 & 52.9 & 87.4 & 95.9 & \textbf{87.4} & 75.5 & 75.9 & 62.3 & \textbf{53.3} & 80.6 & 77.3 & \underline{54.9} & 29.5 & 37.9 & 73.10 & 0.20 & 17.58 \\
NOAH~\cite{zhang2022noah} [arXiv'22] & 69.6 & 92.7 & 70.2 & 99.1 & 90.4 & 86.1 & 53.7 & 84.4 & 95.4 & 83.9 & 75.8 & 82.8 & 68.9 & 49.9 & 81.7 & 81.8 & 48.3 & 32.8 & 44.2 & 73.25 & 0.36 & 17.58 \\
ConvPass~\cite{jie2022convpass} [arXiv'22] & 72.3 & 91.2 & 72.2  & 99.2 & 90.9 & \textbf{91.3} & 54.9 & 84.2 & 96.1 & 85.3 & 75.6 & 82.3 & 67.9 & 51.3 & 80.0 & \textbf{85.9} & 53.1 & \underline{36.4} & 44.4 & 74.45 & 0.33 & 17.64  \\
AdaptFormer~\cite{chen2022adaptformer} [NeurlPS'22]  & 70.8 & 91.2 & 70.5 & 99.1 & 90.9 & 86.6 & 54.8 & 83.0 & 95.8 & 84.4 & \underline{76.3} & 81.9 & 64.3 & 49.3 & 80.3 & 76.3 & 45.7 & 31.7 & 41.1 & 72.32 & 0.16 & 17.61 \\
FacT~\cite{jie2023fact} [AAAI'23] & 71.3 & 89.6 & 70.7 & 98.9 & 91.0 & 87.8 & 54.6 & 85.2 & 95.5 & 83.4 & 75.7 & 82.0 & \underline{69.0} & 49.8 & 80.0 & 79.2 & 48.4 & 34.2 & 41.4 & 73.04 & \underline{0.04} & 17.58 \\
Res-Tuning~\cite{jiang2024res-tuning} [NeurlPS'23] & \underline{75.2} & 92.7 & 71.9 & 99.3 & \underline{91.9} & 86.7 & \underline{58.5} & 86.7 & 95.6 & 85.0 & 74.6 & 80.2 & 63.6 & 50.6 & 80.2 & \underline{85.4} & \textbf{55.7} & 31.9 & 42.0 & 74.09 & 0.51 & 17.67 \\
SynQT~\cite{zhang2024synqt} [ECCV'24] & 70.9 & 89.7 & 68.8 & 98.5 & 89.6 & 77.8 & 50.6 & 82.3 & \textbf{96.7} & 83.5 & 75.2 & 71.8 & 62.7 & 48.5 & 75.4 & 74.1 & 49.0 & 31.7 & 36.1 & 70.15 & 2.73 & 17.20 \\ 
PYRA~\cite{xiong2024pyra} [ECCV'24] & 67.5 & 90.3 & 69.3 & 98.9 & 90.0 & 84.6 & 53.1 & 83.3 & 95.7 & 83.3 & 75.2 & 82.6 & 68.9 & 50.8 & 80.0 & 81.8 & 45.8 & 32.2 & 42.8 & 72.43 & 0.30 & 16.37 \\

DyT $r=0.7$ \cite{zhao2024dyt} [NeurlPS'24] & 74.4 & \underline{95.5} & \underline{73.6} & 99.2 & 91.7 & 87.5 & 57.4 & \underline{88.3} & 96.1 & 86.7 & 76.8 & 83.5 & 63.8 & 52.9 & 83.1 & 85.7 & 54.9 & 34.3 & 45.9 & 75.33 & 0.16 & \underline{14.92} \\
DyT $r=0.9$ \cite{zhao2024dyt} [NeurlPS'24] & 74.3 & 94.9 & 73.1 & 99.2 & 91.4 & 87.8 & 57.1 & 87.9 & 96.1 & 85.9 & 76.0 & 83.3 & 64.8 & 51.5 & 83.4 & 84.0 & 54.8 & 35.1 & 46.4 & 75.11 & 0.16 & 17.07 \\
\midrule

\multicolumn{23}{c}{\emph{The proposed Sparse-Tuning}} \\
\midrule
\rowcolor{tabhighlight}
Sparse-Tuning $r=0.7$ & 74.9 & \textbf{95.7} & 73.5 & \textbf{99.6} & \textbf{92.2} & \underline{90.3} & 57.6 & \textbf{88.7} & 96.3 & \underline{86.9} & 76.9& \underline{83.7} & 63.6 & \underline{52.6} & \underline{83.8} & 82.3& 55.3& \textbf{36.9} & \underline{46.3} & \textbf{75.64} & 0.16 & \textbf{11.65} \\ 
\rowcolor{tabhighlight}
Sparse-Tuning $r=0.9$  & 74.8 & \underline{95.5} & 73.2 & \underline{99.4} &91.7 &88.1 & \textbf{58.7} & 88.2 & \underline{96.4} & 85.8& \textbf{76.4} & \textbf{84.7} & 65.2 & 51.7 & \textbf{84.6} & 84.6 &54.7 &35.2 & \textbf{46.6} & \underline{75.55} & 0.16  & 15.62 \\

\bottomrule

\end{tabular}

\end{table*}

%% file: Tables/CIFAR_100.tex
\begin{table*}[ht]
    \caption{\textbf{Comparison with Mainstream PEFT Methods on CIFAR-100.} This table replicates the data shown in Figure \ref{fig:efficiency}.}
    \label{tab:CIFAR-100}
\addtolength{\tabcolsep}{6pt}
    \centering
     
     \small
     \begin{tabular}{lccccccc}
     \toprule
     \multirow{2}{*}{Method} &  \multirowcell{2}{Params. (M) $\downarrow$} & \multicolumn{2}{c}{Memory Usage (GB)} & \multicolumn{2}{c} {Time (sec/epoch)} & \multirowcell{2}{GFLOPs $\downarrow$} & \multirowcell{2}{Acc.}\\  
     \cmidrule{3-4}
     \cmidrule{5-6}
     & & Train $\downarrow$ & Inference $\downarrow$ & Train $\downarrow$  & Inference $\downarrow$ & \\               
     \midrule

     Full fine-tuning &  85.8   &  17.93   &   2.57   &     90 &  8.03    &  17.58  &  90.91   \\
     \midrule

     Adapter \cite{houlsby2019adapter} &  1.19   &  \underline{13.74}   &   2.61    &   75   &  8.16  &   17.81   & 91.76  \\

     LoRA \cite{hu2021lora} &  2.16   &  14.07   &  2.79    &   72   &  8.04    &   17.58   & 91.42  \\

     AdaptFormer \cite{chen2022adaptformer} & 1.19   & 13.26    &   2.61   &   \underline{68}   &  8.09  &  17.81  & \underline{92.03}  \\
 
     VPT \cite{jia2022vpt} &  \textbf{0.07}   &  14.60   &  2.80    &   77  &   8.24   &  18.32   &   91.64  \\

     DyT \cite{zhao2024dyt} & 4.80   &  14.32  &  \underline{2.15}   &   95  &  \underline{7.12}   &   12.21   & 91.01  \\

     \midrule
     
     \rowcolor{tabhighlight} 
     Sparse-Tuning   & \underline{1.10}    &  \textbf{8.89}   &   \textbf{1.53}   &   \textbf{40}     &   \textbf{4.13}     &  \textbf{11.70}   &   \textbf{92.31}  \\
    
    \bottomrule
    \end{tabular}

\end{table*}

%% file: Tables/sparse_comparisons.tex
\begin{table}[t]
    \caption{\textbf{Comparison with efficient transformers.} ``Params. (M)'' denotes the number of trainable parameters in backbones. The  keeping rate of visual tokens for all methods is 0.5.} 
\label{other tc}
 \addtolength{\tabcolsep}{-6pt}
    \centering
    \scriptsize
        \begin{tabular}{c | c |c |c| c  }
            \toprule
            \multirow{2}{*}{Method } & \multicolumn{1}{c|}{VTAB-1K $\uparrow$} & \multicolumn{1}{c|}{FLOPs (G) $\downarrow$} & \multicolumn{1}{c|}{Param.$\downarrow$} & \multicolumn{1}{c}{Throughput $\uparrow$} \\
             & Accuracy & (G) & (M) & (img/s) \\ 
            \midrule
            
            AdaptFormer \cite{chen2022adaptformer} + DynamicViT \cite{rao2021dynamicvit} & 73.48 & 14.32& 3.1 & 954.82 \\

            AdaptFormer \cite{chen2022adaptformer} + EViT \cite{liang2022evit} & 72.30 & 11.77 & 0.16 & 1152.38 \\
            
            AdaptFormer \cite{chen2022adaptformer} + ToMe \cite{Bolya2023ToMe} & 72.32 & 13.59 & 0.16 & 941.70 \\

            Adapter \cite{houlsby2019adapter} + ToMe \cite{Bolya2023ToMe} & 73.13 & 8.52 & 0.16 & 1180.91 \\
         
            LoRA \cite{hu2021lora} + EViT \cite{liang2022evit} & 72.72 & \textbf{6.99} & 0.16 & \underline{1423.76} \\
             LoRA \cite{hu2021lora} + DynamicViT \cite{rao2021dynamicvit} & 72.38 & 9.02 & 0.16 & 1225.54 \\

            \midrule
            DyT\cite{zhao2024dyt}  & 74.94 & 12.54 & 0.16 & 912.39 \\ 
            DyT\cite{zhao2024dyt} +ToMe \cite{Bolya2023ToMe} & 73.78 & 9.85 & 0.16 & 1114.70 \\
            \midrule
            \rowcolor{tabhighlight}
            Sparse-Tuning (DynamicViT) & 75.19 & 9.15 & 0.16   & 1225.91 \\
            \rowcolor{tabhighlight}
            Sparse-Tuning (EViT)        & \underline{75.63}     & \underline{7.09} & 0.16   & \textbf{1427.28} \\
            \rowcolor{tabhighlight}
            Sparse-Tuning (ToMe)       & \textbf{75.66}  &8.61 & 0.16   &1183.76\\
            \bottomrule
        \end{tabular}

\end{table}

%% file: Tables/ablation.tex
\begin{table}[t]
\caption{\textbf{Ablation on different components of Sparse-Tuning.} Without any component of Sparse-Tuning, it can be viewed as freezing the pre-trained ViT, and only fine-tuning the classification layer.}
\label{Table:TS and DA}
 \addtolength{\tabcolsep}{-5pt}
    \centering
    \scriptsize

\begin{tabular}{l | cc | c | c | c  c  c }
\toprule
\multirow{2}{*}{\#} & \multicolumn{1}{c}{Token}  & \multicolumn{1}{c|}{Dense} & \multirow{2}{*}{Params. (M)$\downarrow$} & \multirow{2}{*}{GFLOPs$\downarrow$} & \multirow{2}{*}{CIFAR-100} & \multirow{2}{*}{SVHN} & \multirow{2}{*}{Food-101} \\
 & Sparsification & Adapters &  &  &  &  &  \\
\midrule

(a) &  &  & \textbf{0} & 17.58 & 85.87 & 56.29 & 88.07 \\ 
\midrule

(b) & \checkmark &  & \textbf{0} & \textbf{10.81} & 76.59 & 48.81 & 78.50 \\
(c) &  & \checkmark & 1.10 & 17.89 & \textbf{92.66} & \textbf{97.93} & \textbf{91.34} \\

\rowcolor{tabhighlight} (d) & \checkmark & \checkmark & 1.10 & 11.70 &  92.31  &  97.47  &  90.72 \\
\bottomrule

\end{tabular}

\end{table}

%% file: Tables/feature_input.tex
\begin{table}[t]
\caption{\textbf{Comparison of different feature inputs.} $x_\text{N}$, $x_{\text{DA}_\text{N-1}}$, and $x_{\text{DA}_\text{N-3}}$ represent the outputs of the MHA at the N-th encoder layer, the outputs of the $\text{DA}_\text{N-1}$ and $\text{DA}_\text{N-3}$.}
\label{Table:feature input}
\addtolength{\tabcolsep}{-4.1pt}
    \centering
    \scriptsize

\begin{tabular}{l | c  c  c  |  c | c |  c  c  c}
\toprule
\# & $x_\text{N}$  &  $x_{\text{DA}_\text{N-1}}$ & $x_{\text{DA}_\text{N-3}}$  & Params. (M)$\downarrow$ & GFLOPs$\downarrow$ & CIFAR-100 & SVHN & Food-101 \\
\midrule

(a) & \checkmark &    &     &  0.69  &  11.19 &  91.10   &  95.67  &  88.73    \\

(b) & \checkmark &  \checkmark  &     &   0.95 &  11.67   &  91.95   &  97.03  &  90.65   \\

(c) & \checkmark &    &  \checkmark   &   0.95  & 11.67  & 91.61  &  96.96  & 90.71   \\

\rowcolor{tabhighlight} (d) & \checkmark &  \checkmark  &  \checkmark   &     1.10    & 11.70 &   \textbf{92.31}  &  \textbf{97.47}  &  \textbf{90.72}   \\

\bottomrule

\end{tabular}

\end{table}

%% file: Tables/feture_fusion.tex
\begin{table}[t]
\caption{\textbf{Comparison of different feature fusion methods.} The positions "Input" and "Output" correspond to (a), (b) in Figure \ref{fig:DA_variants}, respectively. "Inner" position correspond to  Figure \ref{fig:overview}.}
\label{Table:feature fusion}
\addtolength{\tabcolsep}{-1.8pt}
    \centering
    \scriptsize

\begin{tabular}{l | c | c | c | c  c  c }
\toprule
\# & Position & Params. (M)$\downarrow$ & GFLOPs$\downarrow$ & CIFAR-100 & SVHN & Food-101\\
\midrule

(a) & Input &  \textbf{0.69}  &  \textbf{11.19}  &  91.05  &  96.55   &  89.72  \\

\rowcolor{tabhighlight}
(b) & Inner & 1.10 & 11.70 &  \textbf{92.31}  &  \textbf{97.47}  &  \textbf{90.72} \\ 

(c) & Output &   1.68  &  11.81   &  91.27   &  97.14   &   90.19  \\

\bottomrule

\end{tabular}

\end{table}

%% file: Tables/token_drop.tex
\begin{table}[t]
\caption{\textbf{Comparison of different strategies for processing irrelevant tokens.} "Drop" refers to directly removing all irrelevant tokens, while "Argmax" denotes retaining the single irrelevant token with the highest attention score.}
\label{Table:Token drop}
\addtolength{\tabcolsep}{-1.8pt}
    \centering
    \scriptsize

\begin{tabular}{l | c | c | c | c  c  c }
\toprule
\# & Strategy & Params. (M)$\downarrow$ & GFLOPs$\downarrow$ & CIFAR-100 & SVHN & Food-101\\
\midrule

(a) & Drop &  1.10  &  \textbf{11.61}   &  91.22  &   96.89  &  89.63  \\ 

(b) & Argmax & 1.10  &  11.70   &  91.82   &  97.01   &  90.14   \\ 

\rowcolor{tabhighlight}
(c) & Merge & 1.10 & 11.70 &  \textbf{92.31}  &  \textbf{97.47}  &  \textbf{90.72} \\

\bottomrule

\end{tabular}

\end{table}

%% file: Tables/sparse_layer.tex
\begin{table}[t]
\caption{\textbf{Comparison of different positions of Token Sparsification.} For instance, "[4, 7, 10]" represents conducting Token Sparsification in the 4th, 7th, and 10th encoder layers.}
\label{Table:sparsification position}
\addtolength{\tabcolsep}{-1.9pt}
    \centering
    \scriptsize
  
\begin{tabular}{l | c | c | c | c  c  c}
\toprule
\# & Position & Params. (M)$\downarrow$ & GFLOPs$\downarrow$ & CIFAR-100 & SVHN & Food-101 \\
\midrule

(a) & [2, 5, 8] & 1.10 & \textbf{10.78} & 89.77 & 94.65 & 88.71 \\

(b) & [3, 6, 9] & 1.10 & 11.35 & 91.03 & 96.21 & 89.76 \\

\rowcolor{tabhighlight} 
(c) & [4, 7, 10] & 1.10 & 11.70 & \textbf{92.31} & \textbf{97.47} & \textbf{90.72} \\

(d) & [5, 8, 11] & 1.10  &  12.73   &  92.12    &   96.69   &  90.24   \\

(e) & [6, 9, 12] & 1.10   &  13.68   &   91.38   &   96.10   &   89.93 \\

\bottomrule

\end{tabular}

\end{table}

%% file: Tables/neck.tex
\begin{table}[t]
\caption{\textbf{Comparison of different bottleneck dimensions of Dense Adapter.} $d$ represents the bottleneck dimensions.}
\label{Table:neck}
\addtolength{\tabcolsep}{-2.2pt}
\renewcommand{\arraystretch}{0.9}
    \centering
    \scriptsize
    \vspace{2mm}
\begin{tabular}{l | c | c | c | c  c  c | c}
\toprule
\# & $d$ & Params. (M)$\downarrow$ & GFLOPs$\downarrow$ & CIFAR-100 & SVHN & Food-101 & Avg. \\
\midrule

(a) & 8 &  0.29   &  11.60 &  91.03   &  96.31    &  89.55  &  92.30 \\

(b) &   16   &  0.56    &  11.63    &  91.75    &   97.02   &   89.92  &  92.90 \\ 

\rowcolor{tabhighlight} 
(c) & 32 &  1.10 & 11.70 &  \textbf{92.31}  &  \textbf{97.47}  &  \textbf{90.72} &  \textbf{93.50}  \\

(d) & 64 &  2.18  &  11.83   &   92.19   &   96.96   &  90.25   &   93.13   \\ 

(e) & 128 &   4.35  & 12.09    &  92.11  &   96.15 &   90.31   &  92.47   \\

\bottomrule

\end{tabular}

\end{table}

%% file: Tables/boundary.tex
\begin{table}[t]
\caption{\textbf{Comparison of different starting layers of Token Sparsification.} For instance, "3" represents starting to conduct Token Sparsification in the 3-th encoder layers.}
\label{Table:sparsification boundary}
\addtolength{\tabcolsep}{-0.5pt}
    \centering
    \scriptsize
  
\begin{tabular}{l | c | c | c | c  c  c}
\toprule
\# & Layer & Params. (M)$\downarrow$ & GFLOPs$\downarrow$ & CIFAR-100 & SVHN & Food-101 \\
\midrule

(a) & 3 & 1.13 & 11.78 & 92.12 & 96.81 & 90.24 \\

\rowcolor{tabhighlight} 
(b) & 4 & 1.10 & 11.70 & \textbf{92.31} & \textbf{97.47} & \textbf{90.72} \\

(c) & 5 & 1.07 & 11.61 & 91.54 & 97.07 & 90.31 \\

\bottomrule

\end{tabular}

\end{table}

%% file: Tables/interval.tex
\begin{table}[t]
\caption{\textbf{Comparison of different interval of layers for Token Sparsification.} For instance, "2" represents applying Token Sparsification every two encoder layers.}
\label{Table:sparsification interval}
\addtolength{\tabcolsep}{-1.8pt}
    \centering
    \scriptsize
  
\begin{tabular}{l | c | c | c | c  c  c}
\toprule
\# & Interval & Params. (M)$\downarrow$ & GFLOPs$\downarrow$ & CIFAR-100 & SVHN & Food-101 \\
\midrule

(b) & 2 & 1.10 & 11.13 & 91.38 & 96.65 & 90.16 \\

\rowcolor{tabhighlight}
(a) & 3 & 1.10 & 11.70 & \textbf{92.31} & \textbf{97.47} & \textbf{90.72} \\

(c) & 4 &1.10  &14.13  & 91.92 & 97.11 &90.36  \\

\bottomrule

\end{tabular}

\end{table}

%% file: Tables/ViT_L.tex
\begin{table}[!t]
\caption{\textbf{Comparison when scaling up the model size to ViT-L.} $r$ is the keeping rate of the semantic-relevant tokens.}
\label{Table:vit-l}
\addtolength{\tabcolsep}{-4pt}
    \centering
    \scriptsize
    
\begin{tabular}{l | c | c | c  c  c }
\toprule
Method & Params. (M)$\downarrow$ & GFLOPs$\downarrow$ & CIFAR-100 & SVHN & Food-101 \\
\midrule

Full fine-tuning  &  303.3   &  61.60 &  92.05   &  97.44    &  90.62 \\ 
\midrule

DyT $r=0.5$  &  3.17   &  43.79 &  93.49   &  97.38    &  91.49 \\

DyT $r=0.7$     &   3.17   &  51.11  &  93.28    &  97.25    &  91.60   \\ 

DyT $r=0.9$  &   3.17   &  60.05    &  93.44    &  97.23    &  91.59    \\ 
\midrule

Sparse-Tuning $r=0.5$ &   2.93  &   \textbf{30.63}  &  93.56   &    97.31   &    91.46   \\

Sparse-Tuning $r=0.7$ & 2.93 &   40.08   &   \textbf{93.97}   &  \textbf{98.23}   &    91.98   \\ 

Sparse-Tuning $r=0.9$ &  2.93   &   53.78     &     93.45   &  98.15   &   \textbf{92.77}   \\

\bottomrule

\end{tabular}

\end{table}

%% file: Tables/segmentation.tex
\begin{table}[!t]
\caption{\textbf{Performance on segmentation ($r=0.7$).}}
\addtolength{\tabcolsep}{10pt}
\label{tab:seg}

    \centering
    \scriptsize
  
\begin{tabular}{l|cc}
\toprule
 Methods & Object & VOC20 \\
    \midrule
    ProxyCLIP~\cite{lan2024ProxyClip} [ECCV'24] & 39.2 & 83.2 \\
    \midrule
    DyT~\cite{zhao2024dyt} [NeurIPS'24] & 39.9 & 82.7 \\
    Sparse-Tuning (Ours) & \textbf{41.5} & \textbf{84.9} \\

\bottomrule

\end{tabular}

\end{table}

%% file: sec/5_conclusion.tex
\section{Conclusion}
\label{Conclusion}

While current PEFT methods adapt pre-trained ViTs in a parameter-efficient manner, they overlook the high computational and GPU memory costs during
inference. Although token sparsification (TS) techniques offer a way to alleviate computational burdens, they inevitably lead to information loss for significant performance decline. In this work, we propose Sparse-Tuning that achieves both computation and memory efficiency through Dense Adapters (DA). DA aggregates comprehensive token information from shallow layers and integrates it into the retained tokens of deeper layers, thereby minimizing performance degradation. By combining TS techniques with DA, Sparse-Tuning achieves quadratic reduction in both computational and memory overhead while preserving model effectiveness and minimizing information loss. We conducted experiments on the VTAB-1K benchmark and multiple complete image/video datasets to demonstrate that our Sparse-Tuning achieves SOTA performance while significantly improving fine-tuning and inference efficiency.


\section{Future Work}
\label{Future_Work} 

In future work, we will explore the potential of Sparse-Tuning from three perspectives:
First, we will investigate dynamic and adaptive token retention strategies. This involves adjusting the token filtering ratio in real-time based on the complexity of the input content to further balance computational efficiency with information integrity. For example, this would mean retaining more local tokens for images with complex textures while applying greater sparsification to those with simple backgrounds.
Second, we will extend the Sparse-Tuning framework to more complex vision tasks, such as video generation\cite{zouaccelerating} and object tracking\cite{hu2023transformer,hu2024toward}. To address temporal continuity and spatial stereo features, we will optimize the cross-frame and cross-view feature fusion mechanisms within the Dense Adapters.
Third, we will broaden the application of Sparse-Tuning to a wider range of domains. For instance, in resource-constrained scenarios like edge devices, we aim to enable the efficient deployment of large language models and multi-modal large language models, thereby enhancing their multi-task adaptation capabilities.